\documentclass[letterpaper, 10 pt, conference]{ieeeconf}  

\IEEEoverridecommandlockouts                              
\usepackage[utf8]{inputenc}
\usepackage{amsmath,amssymb,stmaryrd,mathtools}

\usepackage{amsfonts}
\usepackage{breqn}

\allowdisplaybreaks

\usepackage{algorithm2e}\RestyleAlgo{ruled}
\usepackage[noend]{algpseudocode}
\usepackage{acronym}
\usepackage{color,xcolor}
\usepackage{verbatim}
\usepackage{booktabs}
\usepackage{siunitx}
\usepackage{graphics}
\usepackage{graphicx,caption}
\usepackage{flushend}
\usepackage{suffix}
\usepackage{xstring}
\usepackage{xparse}
\usepackage{expl3}
\usepackage{mathrsfs}
\usepackage{tabularx}
\usepackage{makecell}
\usepackage{array}
\usepackage{hyperref}
\usepackage{cleveref}
\usepackage{multirow}
\usepackage{diagbox}
\usepackage{rotating}

\usepackage{subcaption}
\usepackage{wrapfig}

\usepackage{tikz}
\usetikzlibrary{arrows,backgrounds,calc}
\usepackage{relsize}
\usepackage{float}
\usepackage{kantlipsum} 
\usepackage{lipsum}
\usepackage{stfloats}
\usepackage{siunitx}

\usepackage[noadjust]{cite}
\usepackage{todonotes}
\usepackage{soul}
\definecolor{smoothgreen}{rgb}{0.7,1,0.7}
\sethlcolor{smoothgreen}

\RequirePackage{luatex85}
\usepackage{pgfplots}
\pgfplotsset{compat=newest}
\pgfplotsset{every axis legend/.append style={%
		cells={anchor=west}}
}
\usepgfplotslibrary{polar}
\usetikzlibrary{arrows}
\tikzset{>=stealth'}

\definecolor{C1}{rgb}{0.0, 0.447, 0.741}
\definecolor{C1_light}{rgb}{0.0, 0.6032388663967612, 1.0}
\definecolor{C2}{rgb}{0.85, 0.325, 0.098}
\definecolor{C3}{rgb}{0.929, 0.694, 0.125}
\definecolor{C4}{rgb}{0.494, 0.184, 0.556}
\definecolor{C5}{rgb}{0.466, 0.674, 0.188}
\definecolor{C6}{rgb}{0.301, 0.745, 0.933}
\definecolor{C7}{rgb}{0.635, 0.078, 0.184}

\usepgfplotslibrary{groupplots}

\usetikzlibrary{shapes.geometric, arrows}

\tikzstyle{startstop} = [rectangle, rounded corners, minimum width=2cm, minimum height=1cm,text centered, draw=black, fill=none]
\tikzstyle{arrow} = [thick,->,>=stealth]

\title{
Learning Task Skills and Goals Simultaneously from Physical Interaction }
\author{Haonan Chen$^*$, Ye-Ji Mun$^*$, Zhe Huang, Yilong Niu, Yiqing Xie,\\ D. Livingston McPherson, Katherine Driggs-Campbell
\thanks{
$^*$ denotes equal contribution. 
}%
\thanks{H. Chen, Y. Mun, Z. Huang, Y. Niu, Y. Xie, D. McPherson, and K. Driggs-Campbell are with the Department of  Electrical and Computer Engineering at the University of Illinois at Urbana-Champaign. emails: \{haonan2, yejimun2, zheh4, yilongn2, yiqingx2, dlivm, krdc\}@illinois.edu}%
\thanks{This work was supported by ZJU-UIUC Joint Research Center for Cyber-physical Manufacturing Networks (CyMaN), Project  No. DREMES 202003, funded by Zhejiang University.}%
}

\renewcommand{\baselinestretch}{0.92}

\begin{document}
\bstctlcite{IEEEexample:BSTcontrol}
\maketitle
\thispagestyle{empty}
\pagestyle{empty}

\begin{abstract}
In real-world human-robot systems, it is essential for a robot to comprehend human objectives and respond accordingly while performing an extended series of motor actions. Although human objective alignment has recently emerged as a promising paradigm in the realm of physical human-robot interaction, its application is typically confined to generating simple motions due to inherent theoretical limitations. In this work, our goal is to develop a general formulation to learn manipulation functional modules and long-term task goals simultaneously from physical human-robot interaction. We show the feasibility of our framework in enabling robots to align their behaviors with the long-term task objectives inferred from human interactions.



\end{abstract}

\section{Introduction}

One of the core challenges in physical human-robot interaction (pHRI) for robotic manipulation is to estimate human goals and adapt the robot's interaction with the environment accordingly~\cite{10160515}. Learning to manipulate objects such as chopping or pouring is relatively easy for a child with parental guidance and feedback, but modeling and planning robot interactions with the environment to do the same can be difficult.
Previous works have explored a variety of strategies for handling pHRI, including generating desired impedance, switching to gravity compensation to comply with human-applied force, or updating the objective function based on real-time interaction~\cite{pmlr-v78-bajcsy17a, doi:10.1177/02783649211050958}. 

\begin{figure}[tbp]
\centering
\includegraphics[width=0.2\textwidth]{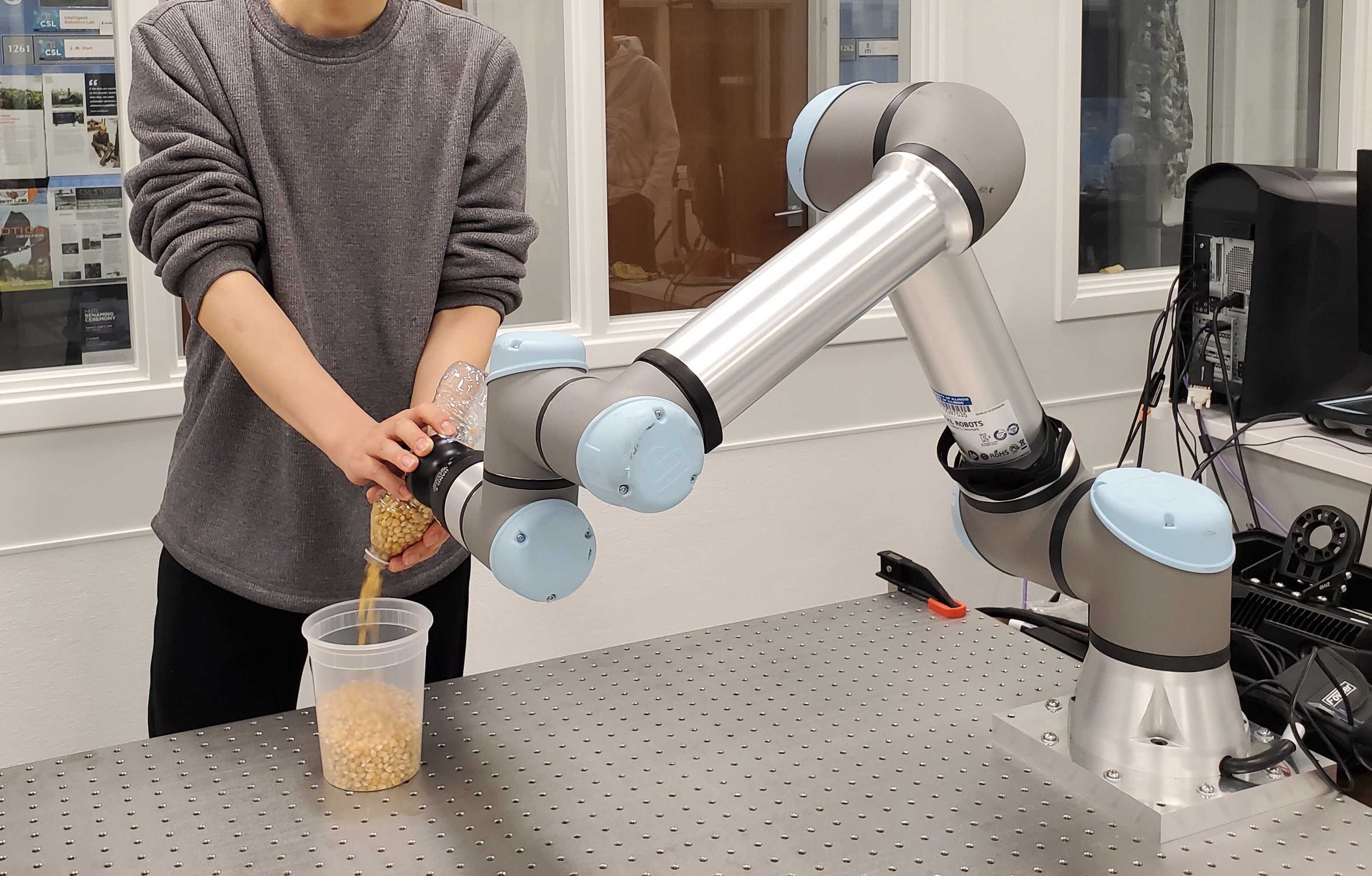}
\caption{Experimental Setup. The participant grasps the robot's end effector to perform the pouring actions, while the robot learns about the desired pouring amount and the pouring skills.}
\label{fig:setup}
\vspace{-20pt}
\end{figure}

These underlying approaches are accompanied by several limitations, such as restricting simple motion generation and lacking the capacity to synthesize intricate motions. In contrast, we introduce behavior primitives and propose a framework that allows robots to learn from human interaction while manipulating liquid or granular materials (see Fig. \ref{fig:setup}). By employing a parameterized action space, the autonomous agent can infer human intent to interact with the object and environment. The incorporation of behavior primitives enables the robot to generate complex behaviors, thereby facilitating operation in more general settings.

In this work, we propose a novel framework that identifies task goals and subsequently update the robot's behavior during interactions with the external environment. We aim to minimize human efforts (i.e., interaction time) to teach the robot to complete the task. To this end, we take a hierarchical approach to decompose the task into high-level task skills (also referred to as behavior primitives in this paper) and low-level parameters for each skill, which allows the robot to learn complex tasks such as pouring. Through the employment of hierarchical modeling, we allow robots to reject human disturbances, estimate high-level controller types, and infer parameters for low-level controllers. We employ a Bayesian inference framework to infer both the desired skills (\textit{e.g.}, shaking, tapping, and stopping) and the long-term task goal (\textit{e.g.}, pouring amount).  






\section{Methodology}
\label{sec:methods}

\subsection{Problem Formulation}
We model pHRI as a discrete system with forward dynamics function $f$ in line with prior works~\cite{9007490, doi:10.1177/02783649211050958, pmlr-v78-bajcsy17a}:
\begin{equation}\label{eq:dynamics}
x_r^{t+1} = f (x_r^t,u_r^t+u_h^t).
\end{equation}
where $x\in \mathcal{R}^{n\times6}$ denotes the joint positions and velocities of the $n$-DOF robot, $u_r \in \mathcal{R}^6$ is commanded velocity of the end effector, and $u_h\in \mathcal{R}^6$ is the velocity of the end effector resulting from the wrench applied by the human at the time step $t$. In the presence of human actions, the robot's trajectory is subject to deformation to conform to human corrections.



We assume that there is a task parameter $\beta$, which captures the desired goal of humans in carrying out the task. The robot, lacking knowledge of the human's true target, relies on human interactions to gain information about this objective. For example, $\beta$ could denote the desired quantity of liquids or powders to complete the pouring task, or the target size for chopping. The robot estimates $\beta$ by filtering based on observations $o^{0:t} = \{x_r^t, x_e^t, u_r^t, u_h^t\}$, where $x_e^t$ represents the state measurement of the environment. In our task setting, $x_e^t$ denotes the measured poured amount. The robot updates a belief $b^t(\beta) =  P(\beta|o^{0:t})$ from the previous history of observations $o^{0:t}$. Utilizing Bayes' theorem, we have:
\begin{equation}\label{eq: bayes}
    P(\beta|o^{0:t}) \propto P(o^{t}|\beta, o^{0:t-1})\,P(\beta|o^{0:t-1})
\end{equation}
We can expand the likelihood $P(o^{t}|\beta, o^{0:t-1})$ as:
\begin{equation}\label{eq:belief-update}
    \begin{aligned}
        P(o^{t}|\beta, o^{0:t-1}) &= P(x_r^{t}, x_e^{t}, u_r^{t}, u_h^{t} | \beta, o^{0:t-1}) \\
        & = P(u_r^{t}| \beta, o^{0:t-1}, x_r^{t}, x_e^{t}, u_h^{t}) \\
         & P(u_h^{t} | \beta, o^{0:t-1}, x_r^{t}, x_e^{t}) 
          P(x_r^{t}, x_e^{t} | \beta, o^{0:t-1}) 
    \end{aligned}
\end{equation}

We make three reasonable assumptions to justify the simplification of the likelihood $P(o^{t}|\beta, o^{0:t-1})$. The first assumption is that the robot's action complies with the human during the interaction while being a deterministic mapping from $\beta$ and $x_e^{t}$ when there is no interaction. Thus, $P(u_r^{t} | \beta, o^{0:t-1}, x_r^{t}, x_e^{t}, u_h^t)$ can be reduced in the equations~\ref{eq:belief-update}. The second assumption is human's action follows the \textit{Markov property}, so $u_h^{t}$ is \textit{independent} of $o^{0:t-1}$ conditioned on the task goal $\beta$ and states $x_r^{t+1}, x_e^{t+1}$. The third assumption is that state transition dynamics are deterministic for both robot and environment, so $P(x_r^{t+1}, x_e^{t+1} | \beta, o^{1:t})$ disappears. We can then express the likelihood as:
\begin{equation}\label{eq: likelihood}
    \begin{aligned}
        P(o^{t}|\beta, o^{0:t-1}) \propto P(u_h^{t} | \beta, x_r^{t}, x_e^{t}) 
    \end{aligned}
\end{equation}

Combining Equation~\ref{eq: bayes} and Equation~\ref{eq: likelihood}, we can now get the iterative posterior distribution over the task goal belief:
\begin{equation}
    b^{t}(\beta) \propto P(u_h^{t} | \beta, x_r^{t}, x_e^{t}) \,b^{t-1}(\beta)
\end{equation}

\subsection{Approximate Inference over Task Goals}
\textbf{Observation Model.}
We model that humans' actions reflect the difference between the current task progress and the desired task goal. For example, in the task of robot pouring, the human operator adjusts the robot's pouring speed based on the difference between the desired and actual amount of liquid. When the gap is large, the operator will adjust the robot to pour more aggressively, while for a smaller gap, the operator will adjust the robot to pour more conservatively. To formalize this relationship, we define a distance function $\Delta(x_e^t, \beta)$ to measure the discrepancy  between the current progress $x_e^t$ and the desired task goal $\beta$. We then use a function $g$ that maps the distance to a probability distribution over the space of possible actions, giving us:
\begin{equation}\label{eq:obs}
p(u_h^t |\beta, x_r^t, x_e^t) \propto p(u_h^t |\beta, x_e^t)\propto g(\Delta(x_e^t, \beta)).
\end{equation}

In the case of robot pouring, the function $g$ can be chosen to be a sigmoid function that maps the discrepancy to a probability of choosing a certain pouring speed, with higher discrepancy values corresponding to higher probabilities of aggressive pouring. In the event that the robot state $x_r^t$ aligns with human expectations, it is expected that humans will refrain from taking action, denoted by $u_h^t$ being zero.

\textbf{Approximating Task Goal Posterior.}
Since there is no linear relationship between the observation model and the task goal, and given the cardinality of the task goal space $|\mathcal{B}|$, we represent the belief as the density of a sampled distribution:
\begin{equation}\label{eq: sampling}
b^t(\beta) = \sum_{i=1}^{|\mathcal{B}|} w_i^t \delta(\beta_i)
\end{equation}
where $\delta(\beta_i)$ is a delta function centered at $\beta_i$. Using \textit{importance sampling}, we can approximate a target distribution $b^t(\beta)$ by drawing samples from a proposal distribution $q(\beta^t|u_h^{0:t})$~\cite{mpc_book}. The weight $w_i^t$ can be represented as:
\begin{align}
    w_i^t & \propto \frac{p(\beta_i|o^{0:t})}{q(b^t(\beta_i)|o^{0:t})}.
\end{align}
The weight can be written recursively as:
\begin{align}
w_i^t \propto w_i^{t-1} \frac{P(u_h^{t} | \beta, x_r^{t}, x_e^{t}) }{q(b^t(\beta_i)|b^{t-1}(\beta_i) , o^{t})}.
\end{align}
Taking the proposal distribution $q(b^t(\beta_i)|b^{t-1}(\beta_i) , o^{t})$ to be a deterministic value, we have:
\begin{align}\label{eq: weight}
      w_i^t  \propto w_i^{t-1}P(u_h^{t} | \beta, x_r^{t}, x_e^{t}).
\end{align}
In practice, we want to ensure the weights are normalized, specifically to satisfy the condition of $\sum_{i=0}^{|B|} w_i^t = 1$. The full algorithm is summarized in Algorithm \ref{alg: goal}.

\begin{algorithm}[tbp]
\caption{Online Goal Learning from pHRI}\label{alg: goal}
    Initialize: $w_{i=0:|\mathcal{B}|}^{t=0} \gets \frac{1}{|\mathcal{B}|}$\\
    \For{$t=0$ to $T$}{
        $u_r^t = B_r(\dot q_r^t - \dot q^t) + K_r(q_r^t - q^t)$ \\
        $\eta=0$ \\
        \uIf{$u_h^t = 0$}{
            $P(u_h^{t}|\beta_i, x_r^{t}, x_e^{t}) \gets 1$   
        }
        \Else{
            $P(u_h^{t}|\beta_i, x_r^{t}, x_e^{t}) \gets P_{g(\Delta(x_e^{t}, \beta_i))}(u_h^{t})$ \Comment{(\ref{eq:obs})}\\
        }
        \For{$i=1$ to $|\mathcal{B}|$}{
        $w_i^{t} \gets w_i^{t-1}p(u_h^t |\beta, x_e^t)$ \Comment{(\ref{eq: weight})}\\
        $\eta \gets \eta + w_i^t$ \\
        }
        \For{$i=1$ to $|\mathcal{B}|$}{
        $w_i^{t} \gets w_i^{t}/\eta$ \\
        }
        $u^t_r \gets Opt(\Delta(\beta_{cur}, b^t(\beta)))$
    }
\vspace{-0pt}
\end{algorithm}

\section{Conclusion and Future Works}

\label{sec:conclusion}
In this work, we introduce a novel framework that employs importance sampling within a Bayesian paradigm to minimize the human effort required in various daily scenarios that involve pHRI. We formalize how the robot can effectively infer task objectives (i.e., target pouring amount) and optimize the task skills (i.e., shaking, pouring, stopping) during the physical interaction. We show the analysis that how task goals can be inferred in complex daily manipulation tasks. We plan to conduct a user study to demonstrate the applicability of the proposed framework in a series of complex pouring tasks involving shaking and tapping motions. In the future, we also intend to evaluate the potential of our proposed approach in terms of its capability to generalize and its feasibility in practice with respect to a range of source containers and pouring materials (e.g., rice, beans, candies, cereals, and carrots).

\bibliographystyle{IEEEtran}
\bibliography{BibFile}

\begin{thebibliography}{1}
\providecommand{\url}[1]{#1}
\csname url@rmstyle\endcsname
\providecommand{\newblock}{\relax}
\providecommand{\bibinfo}[2]{#2}
\providecommand\BIBentrySTDinterwordspacing{\spaceskip=0pt\relax}
\providecommand\BIBentryALTinterwordstretchfactor{4}
\providecommand\BIBentryALTinterwordspacing{\spaceskip=\fontdimen2\font plus
\BIBentryALTinterwordstretchfactor\fontdimen3\font minus
  \fontdimen4\font\relax}
\providecommand\BIBforeignlanguage[2]{{%
\expandafter\ifx\csname l@#1\endcsname\relax
\typeout{** WARNING: IEEEtran.bst: No hyphenation pattern has been}%
\typeout{** loaded for the language `#1'. Using the pattern for}%
\typeout{** the default language instead.}%
\else
\language=\csname l@#1\endcsname
\fi
#2}}

\bibitem{10160515}
Z.~Huang, Y.-J. Mun, X.~Li, Y.~Xie, N.~Zhong, W.~Liang, J.~Geng, T.~Chen, and
  K.~Driggs-Campbell, ``Hierarchical intention tracking for robust human-robot
  collaboration in industrial assembly tasks,'' in \emph{2023 IEEE
  International Conference on Robotics and Automation (ICRA)}, 2023, pp.
  9821--9828.

\bibitem{pmlr-v78-bajcsy17a}
\BIBentryALTinterwordspacing
A.~Bajcsy, D.~P. Losey, M.~K. O’Malley, and A.~D. Dragan, ``Learning robot
  objectives from physical human interaction,'' in \emph{Proceedings of the 1st
  Annual Conference on Robot Learning}, ser. Proceedings of Machine Learning
  Research, S.~Levine, V.~Vanhoucke, and K.~Goldberg, Eds., vol.~78.\hskip 1em
  plus 0.5em minus 0.4em\relax PMLR, 13--15 Nov 2017, pp. 217--226. [Online].
  Available: \url{https://proceedings.mlr.press/v78/bajcsy17a.html}
\BIBentrySTDinterwordspacing

\bibitem{doi:10.1177/02783649211050958}
\BIBentryALTinterwordspacing
D.~P. Losey, A.~Bajcsy, M.~K. O’Malley, and A.~D. Dragan, ``Physical
  interaction as communication: Learning robot objectives online from human
  corrections,'' \emph{The International Journal of Robotics Research},
  vol.~41, no.~1, pp. 20--44, 2022. [Online]. Available:
  \url{https://doi.org/10.1177/02783649211050958}
\BIBentrySTDinterwordspacing

\bibitem{9007490}
A.~Bobu, A.~Bajcsy, J.~F. Fisac, S.~Deglurkar, and A.~D. Dragan, ``Quantifying
  hypothesis space misspecification in learning from human–robot
  demonstrations and physical corrections,'' \emph{IEEE Transactions on
  Robotics}, vol.~36, no.~3, pp. 835--854, 2020.

\bibitem{mpc_book}
J.~Rawlings, D.~Mayne, and M.~Diehl, \emph{Model Predictive Control: Theory,
  Computation, and Design}, 01 2017.

\end{thebibliography}
\clearpage
\end{document}